\UseRawInputEncoding
\documentclass[letterpaper, 10 pt, conference]{ieeeconf}  % Comment this line out if you need a4paper

\IEEEoverridecommandlockouts                              % This command is only needed if 
\usepackage{amsmath,amssymb,amsopn,amstext,amsfonts}
\usepackage{algorithmic}
\usepackage{array}

\usepackage{graphicx}
\usepackage{cite}

\usepackage{tabu}                     % Insertion of table
\usepackage{multirow}                 % It is generally used for designing tables and merging rows.
\usepackage{multicol}                 % Merging multiple columns
\usepackage{float}                    % Floating images
\usepackage{makecell}                 % Three-line table - vertical line
\usepackage{subcaption}               % For subfigure environment
\usepackage{booktabs}                 % Three-line table - short thin horizontal line
\usepackage{balance}
\title{\LARGE \bf
Diffusion-Based mmWave Radar Point Cloud Enhancement \\ Driven by Range Images
}
\author{Ruixin Wu$^{*1,2,3}$, Zihan Li$^{*4}$, Jin Wang$^\dag$$^{1,2,3}$, Xiangyu Xu$^{1,2,3}$, Zhi Zheng$^{1,2,3}$, \\Kaixiang Huang$^{1,2,3}$ and Guodong Lu$^{1,2,3}$
\thanks{ $^*$Indicates equal contribution. }
\thanks{$^\dagger$Corresponding author: {\tt\small dwjcom@zju.edu.cn}.}
\thanks{
	$^{1}$The State Key Laboratory of Fluid Power and Mechatronic Systems, School of Mechanical Engineering, Zhejiang University, Hangzhou 310058, China. $^{2}$Zhejiang Key Laboratory of Industrial Big Data and Robot Intelligent Systems, Zhejiang University, Hangzhou 310058, China. $^{3}$Robotics Research Center of Yuyao City, Ningbo 315400, China. $^{4}$State Key Laboratory of Robotics and Systems, Department of Mechatronics Engineering, Harbin Institute of Technology, Harbin 150001, China. }
    }
    
\begin{document}
\maketitle 
\thispagestyle{empty}
\pagestyle{empty}

%%%%%%%%%%%%%%%%%%%%%%%%%%%%%%%%%%%%%%%%%%%%%%%%%%%%%%%%%%%%%%%%%%%%%%%%%%%%%%%%
\begin{abstract}
Millimeter-wave (mmWave) radar has attracted significant attention in robotics and autonomous driving. However, despite the perception stability in harsh environments, the point cloud generated by mmWave radar is relatively sparse while containing significant noise, which limits its further development. Traditional mmWave radar enhancement approaches often struggle to leverage the effectiveness of diffusion models in super-resolution, largely due to the unnatural range-azimuth heatmap (RAH) or bird's eye view (BEV) representation. To overcome this limitation, we propose a novel method that pioneers the application of fusing range images with image diffusion models, achieving accurate and dense mmWave radar point clouds that are similar to LiDAR. Benefitting from the projection that aligns with human observation, the range image representation of mmWave radar is close to natural images, allowing the knowledge from pre-trained image diffusion models to be effectively transferred, significantly improving the overall performance. Extensive evaluations on both public datasets and self-constructed datasets demonstrate that our approach provides substantial improvements, establishing a new state-of-the-art performance in generating truly three-dimensional LiDAR-like point clouds via mmWave radar. Code will be released after publication.
\end{abstract}

%%%%%%%%%%%%%%%%%%%%%%%%%%%%%%%%%%%%%%%%%%%%%%%%%%%%%%%%%%%%%%%%%%%%%%%%%%%%%%%%
\section{Introduction}
\label{sec:Introduction}
In recent years, mmWave radar has found increasing applications in tasks such as Advanced Driver Assistance Systems (ADAS) and Navigation-Assisted Driving (NOA). Moreover, the application of mmWave radar has also expanded to various robotic systems, including mobile robots \cite{lyu2024robust}, unmanned aerial vehicles (UAVs) \cite{zhang20234dradarslam}, and unmanned surface vehicles (USVs) \cite{guan2023achelous}, \cite{cheng2021robust}. MmWave radar, with its compact structure and robust performance, remains reliable in extreme weather such as rain, snow, and fog, making it an ideal choice for environmental perception in harsh environments.

Despite the obvious advantages of mmWave radar, it is typically used as a supplementary sensor in practical applications, providing supporting information rather than being used independently. This situation can be attributed to two primary factors:

\begin{enumerate}
\item{{\bf Cluttered: }In a challenging environment, electromagnetic wave propagation varies significantly due to phenomena such as reflection, scattering, refraction, and diffraction. Furthermore, radiated energy during transmission and reception can create sidelobe signals. These combined effects can lead to the formation of a large number of false point clouds and false targets (ghost points).}
\item{{\bf Sparse: }Compared to LiDAR, radar point clouds are highly sparse, lacking sufficient geometric and detailed information, thus limiting their effectiveness in high-precision tasks like detection, localization, and mapping.}
\end{enumerate}

\begin{figure}[t]
	\centering
        \includegraphics[width=\linewidth]{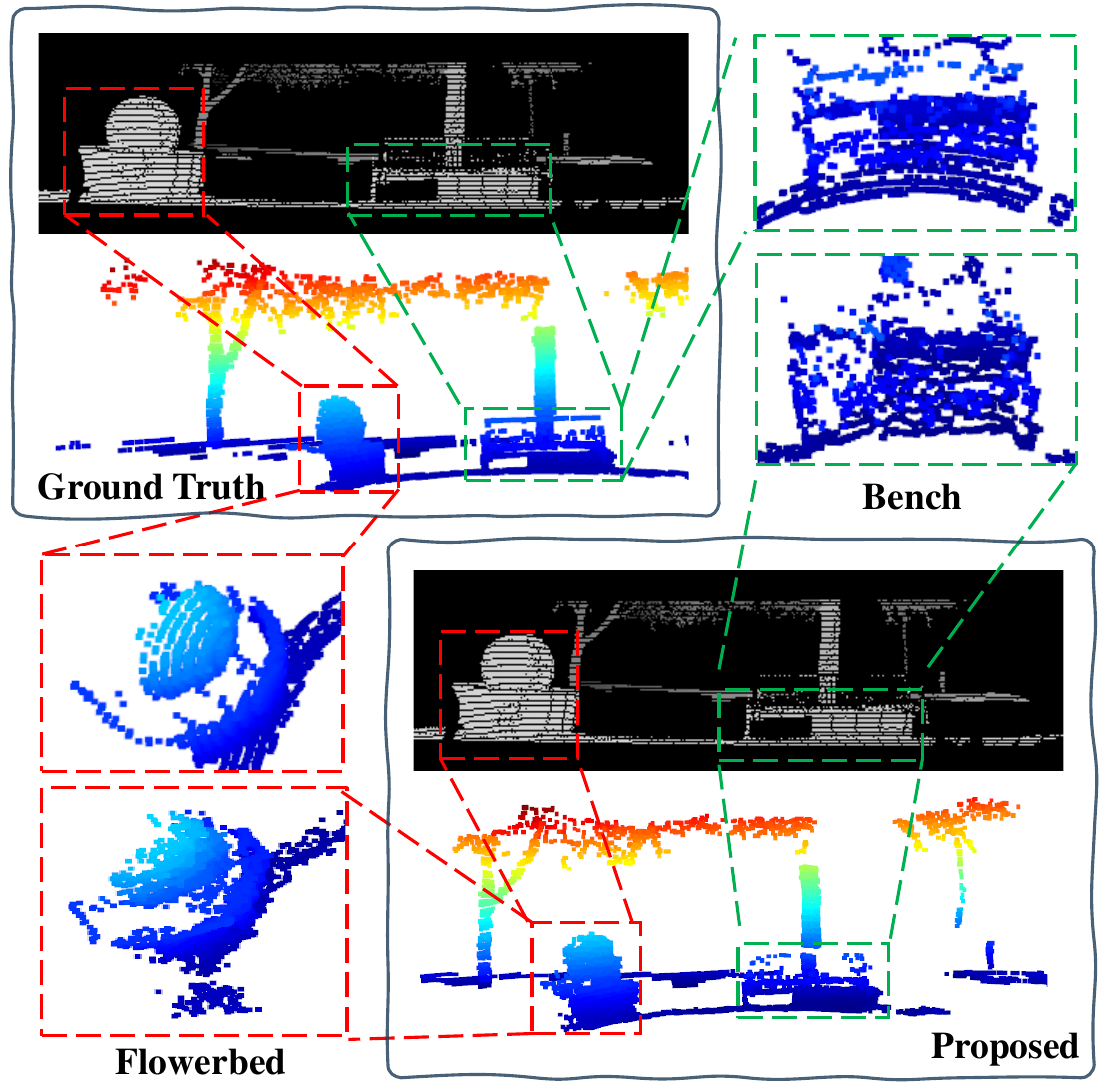}
        \caption{The comparative demonstration includes LiDAR range images and point clouds, as well as the range images and high-quality millimeter-wave radar point clouds obtained through the proposed method. The red and green dashed boxes highlight the flowerbed and bench in the current scene, respectively. Humans can easily identify object categories and geometric structures through range images, demonstrating that range images preserve rich semantic and geometric information. As a proxy representation of point clouds, it enables the knowledge of pre-trained image diffusion models to be fully leveraged in point cloud super-resolution tasks.}
	\label{selfmade}
\end{figure}

To address these limitations, extensive research has focused on two aspects, corresponding to the two stages in the mmWave radar signal processing pipeline: First, improving pre-processing methods to enhance the angular resolution of the point cloud, reduce false detections, and improve data quality \cite{cheng2022novel, prabhakara2023high, zhang2024towards}. Second, post-processing enhances the geometric and detailed information of the point cloud while reducing the workload on the detector. The point clouds obtained by existing methods still have poor accuracy and density, limiting their suitability for backend tasks in various robotic systems. Therefore, the enhancement of mmWave radar remains an open problem.

Since the requirement of generating high-quality mmWave radar point clouds from raw information is to reduce noise and fill sparse areas, the mmWave radar enhancement task can be regarded as a super-resolution task. Thus, the application of diffusion models to mmWave radar is quite natural. 

At present, several related studies \cite{zhang2024towards, Luan2024Diffusion, wu2024diffradar} have explored the application of diffusion models to the generation of mmWave radar point clouds. Nearly all of these approaches employ mmWave radar data, structured in a particular format, as a conditioning input to guide the diffusion process.
For example, Zhang et al. \cite{zhang2024towards} apply RAH as the 2D representation of mmWave radar and successfully generate BEV maps of LiDAR-like quality. Some studies \cite{Luan2024Diffusion, wu2024diffradar} have also explored converting raw mmWave radar point clouds into BEV to steer diffusion models in the generation of dense point clouds.
These methods, including RAH and BEV, emphasize velocity information and top-down perspectives, which, unlike natural images, lack abundant pixel-level structural details and semantic context. However, diffusion models customarily depend on such information to achieve effective representation learning. Consequently, there is an urgent need for a more suitable representation of mmWave radar data to enable diffusion models to adapt effectively to these inputs.

To overcome this problem, we propose a novel approach that, for the first time, integrates range images with diffusion models to generate mmWave radar point clouds. Unlike other representations, range images preserve more geometric and semantic information while aligning better with human perceptual patterns and natural images, which can be intuitively observed in Fig. \ref{selfmade}. This enables the diffusion model to seamlessly adapt to mmWave radar data and perform representation learning with greater efficiency, thereby markedly enhancing the performance of mmWave radar reconstruction.

To validate the performance of our approach, we conducted extensive benchmark comparisons with existing methods on public datasets and seamlessly transferred the models trained on these datasets to our self-constructed datasets. The results indicate that our method markedly surpasses existing techniques in mmWave radar point cloud quality and exhibits some generalization capability, demonstrating its superior performance.

In summary, the main contributions of this paper are as follows:
\begin{itemize}
    \item [1)] 
    We propose a novel high-quality mmWave radar point cloud generation method based on a pre-trained diffusion model that can generate truly three-dimensional dense and accurate LiDAR-like point clouds.
    \item [2)]
    Our work pioneers the integration of range image representations with diffusion models for mmWave radar super-resolution. By introducing human perception-aligned range images as a proxy representation of point clouds, we can fully leverage the prior knowledge embedded in pre-trained models.
    \item [3)]
    We conducted extensive benchmark, generalization and ablation experiments on both public and self-constructed datasets, which validated the superior performance of the proposed approach.
\end{itemize}	

\section{Related Work}
Traditional detection methods, such as constant false alarm rate (CFAR) \cite{barkat1989cfar,minkler1990cfar,gandhi1988analysis,rohling1983radar} and multiple signal classification (MUSIC) \cite{Schmidt1986Multiple}, generate point clouds that are either excessively sparse or have low signal-to-noise ratios. Existing research on improving the quality of mmWave radar point clouds primarily focuses on improving pre-processing methods and post-processing methods. 

\subsection{Pre-processing Methods}
Cheng et al. \cite{cheng2022novel} generated ground truth Range-Doppler Maps (RDM) from LiDAR point clouds and used them to train Generative Adversarial Networks (GANs) for building detectors. While this approach improves detection performance over traditional methods, it still relies on traditional DOA estimation, leading to noisy mmWave radar point clouds. 
Prabhakara et al. \cite{prabhakara2023high} supervised a U-Net network using LiDAR point cloud labels, with low-resolution radar 2D heatmaps as input, to directly generate LiDAR-like point clouds. This approach retains real object data in the scene while eliminating part of the noise.
Han et al. \cite{han2024denserradar} introduced the DenserRadar network, which directly processes the massive raw mmWave radar 4D cube using a three-dimensional U-Net architecture. However, its computational overhead is enormous.
Zhang et al. \cite{zhang2024towards} supervised a diffusion model using LiDAR BEV images, which predicts LiDAR-like BEV images from paired radar RAH. 
The methods mentioned above use neural networks' cross-modal learning to directly generate more accurate and dense point cloud data without DOA estimation. 
In addition to approaches founded on LiDAR supervision, there are also methods that employ visual-inertial systems to produce ground truth labels for the purpose of cross-modal supervision. For example, Fan et al. \cite{fan2024enhancing} generated labels using a dynamic 3D reconstruction algorithm, thereby replacing the LiDAR-based labels in RPDNet and attaining performance on par with LiDAR-based supervision.
However, these methods either fall short of producing authentic 3D point clouds or yield point clouds that remain comparatively sparse, thereby constraining their suitability for more advanced applications, such as autonomous navigation within complex environments.

\subsection{Post-processing Methods}
Lu et al. \cite{lu2020see} proposed MilliMap, which uses a GAN and takes sparse maps obtained from Bayesian grid mapping as input. This method can reconstruct dense grid maps. 
Geng et al. \cite{geng2023dream} first applied non-coherent integration and synthetic aperture accumulation methods to improve the density and angular resolution of radar point clouds, and then proposed the RDM network, which further filters out noise under the guidance of LiDAR point clouds. However, both methods rely on accurate ego-motion estimation from other sensors, which is difficult to obtain in extreme environments.
Cai et al. \cite{cai2023millipcd} combined traditional signal processing with an adaptive neural network to generate high-quality indoor point clouds from mmWave reflection signals. Although this method integrates mmWave data collected from multiple devices, the resulting point clouds still lack sufficient geometric shape and detailed information.
Wu et al. \cite{wu2024diffradar} projected the raw mmWave radar point cloud into a BEV representation and extracted features as conditioning inputs to guide the diffusion model in reconstructing the point cloud. However, this data structure lacks abundant structural and semantic information, preventing the diffusion model from performing efficient representation learning.

In summary, existing methods either require fusion with other sensors to overcome the limitations of mmWave radar, or produce point clouds that are still not dense or accurate enough, making it challenging to support autonomous navigation or other advanced tasks in complex environments

\section{Preliminaries}
\label{sec: PRELIMINARIES}
\begin{figure*}[t] 
    \centering
    \includegraphics[width=\linewidth]{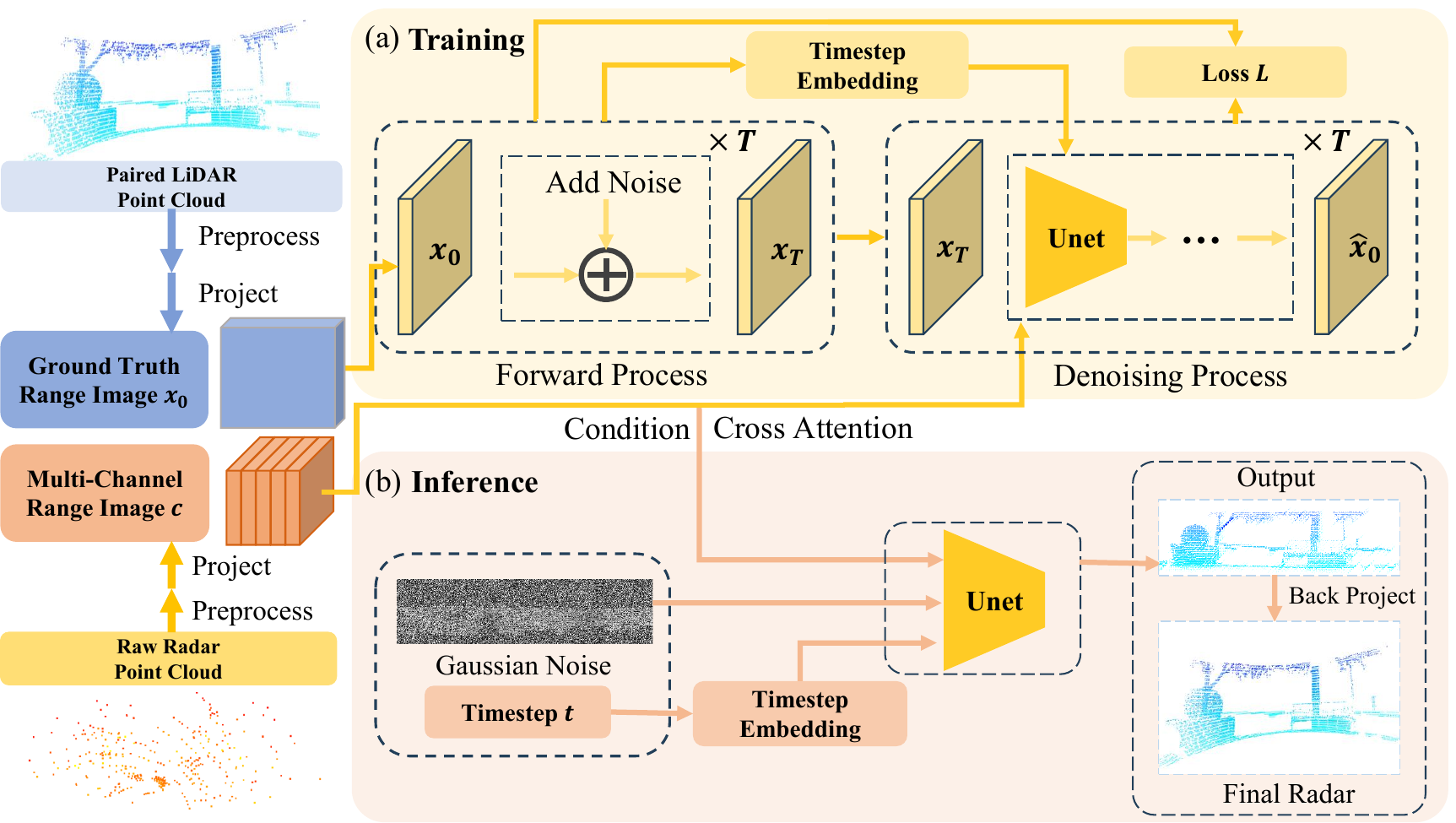}
    \caption{Diagram of the architecture of our proposed approach. (a) In training, the ground truth LiDAR point cloud and the raw mmWave radar point cloud are respectively projected into the range image $x_0$ and the multi-channel range image $c$. Then $x_0$ is corrupted to $x_T$ through the forward process. Finally, the neural network is trained to estimate the ground truth range image, conditioned on the paired radar multi-channel range images. (b) In inference, the neural network directly predicts $\hat{x}_0$ from pure Gaussian noise conditioned on the paired radar multi-channel range images. Subsequently, the predicted range image $\hat{x}_0$ is back-projected to obtain the final high-quality mmWave radar point cloud.}
    \label{system}
\end{figure*}

\subsection{Diffusion Models}
\label{diffusion}
Diffusion models can be encapsulated within a unified generative modeling framework proposed by Karras et al. \cite{karras2022elucidating}. They conceptualize the diffusion process, which is represented as a stochastic differential equation (SDE):
\begin{equation}
	\label{diffusion1}
        dx = f(t) x \, dt + g(t) \, d\omega_t,
\end{equation} 
where $\omega_t$ denotes the standard Wiener process, $f(\cdot):\mathbb{R} \to \mathbb{R}$ and $g(\cdot):\mathbb{R} \to \mathbb{R}$ represent the drift and diffusion coefficients, respectively, with $d$ indicating the dimensionality of the dataset.

The diffusion model learns to reverse a forward diffusion process, which corresponds to the forward SDE in Eq. \ref{diffusion1} and is defined as follows:

\begin{equation}
	\label{diffusion2}
	q(x_t | x_{0}) := \mathcal{N}( x_t; s(t)x_{0}, s^2(t) \sigma^2(t) \mathcal{I} ),
\end{equation}

where $q(\cdot | \cdot)$ represents the conditional probability, $\mathcal{N}(x; \mu, \Sigma)$ denotes the probability density function of $\mathcal{N}(\mu, \Sigma)$ evaluated at $x$, $t \in \{1, \dots, T\}$, where $T$ is the total number of steps in the diffusion chain, $s(t) = \exp \left( \int_0^t f(\xi)\, d\xi\right)$, and $\sigma(t)=\sqrt{ \int_0^t \frac{g(\xi)^2}{s(\xi)^2}\, d\xi}$. Gaussian noise can be added to $x_0$ to yield any $x_{t}$ in a single step. The forward process models a fixed Markov chain, and the noise depends on a variance schedule $s^2(t) \sigma^2(t)$.

The reverse process is defined by the following formula, with parameters $\theta$:

\begin{equation}
	\label{diffusion3}
	p_\theta(x_0) := \int p_\theta(x_{0:T}) \, dx_{1:T},
\end{equation} 
\begin{equation}
	\label{diffusion4}
	p_\theta(x_{0:T}) = p(x_T) \prod_{t=1}^{T} p_\theta(x_{t-1}|x_t),
\end{equation} 
\begin{equation}
	\label{diffusion5}
	p_\theta(x_{t-1}|x_t) := \mathcal{N}(x_{t-1}; \mu_\theta(x_t, t), \Sigma_\theta(x_t, t)).
\end{equation} 
By employing a fixed variance $\Sigma_\theta(x_t, t)$, our task reduces to learning the means of the reverse process, $\mu_\theta(x_t, t)$. Training is conventionally carried out through a reweighted variational bound on the maximum likelihood objective, with the loss defined as follows:
\begin{equation}
	\label{diffusion6}
        L := \mathbb{E}_{t,q} \left[ \lambda_t \left\| \mu_t(x_t, x_0) - \mu_\theta(x_t, t) \right\|^2 \right],
\end{equation} 
where $\mu_t(x_t, x_0)$ is the mean of the forward process posterior $q(x_{t-1} | x_t, x_0)$.

\begin{figure}[t] 
	\centering
        \includegraphics[width=\linewidth]{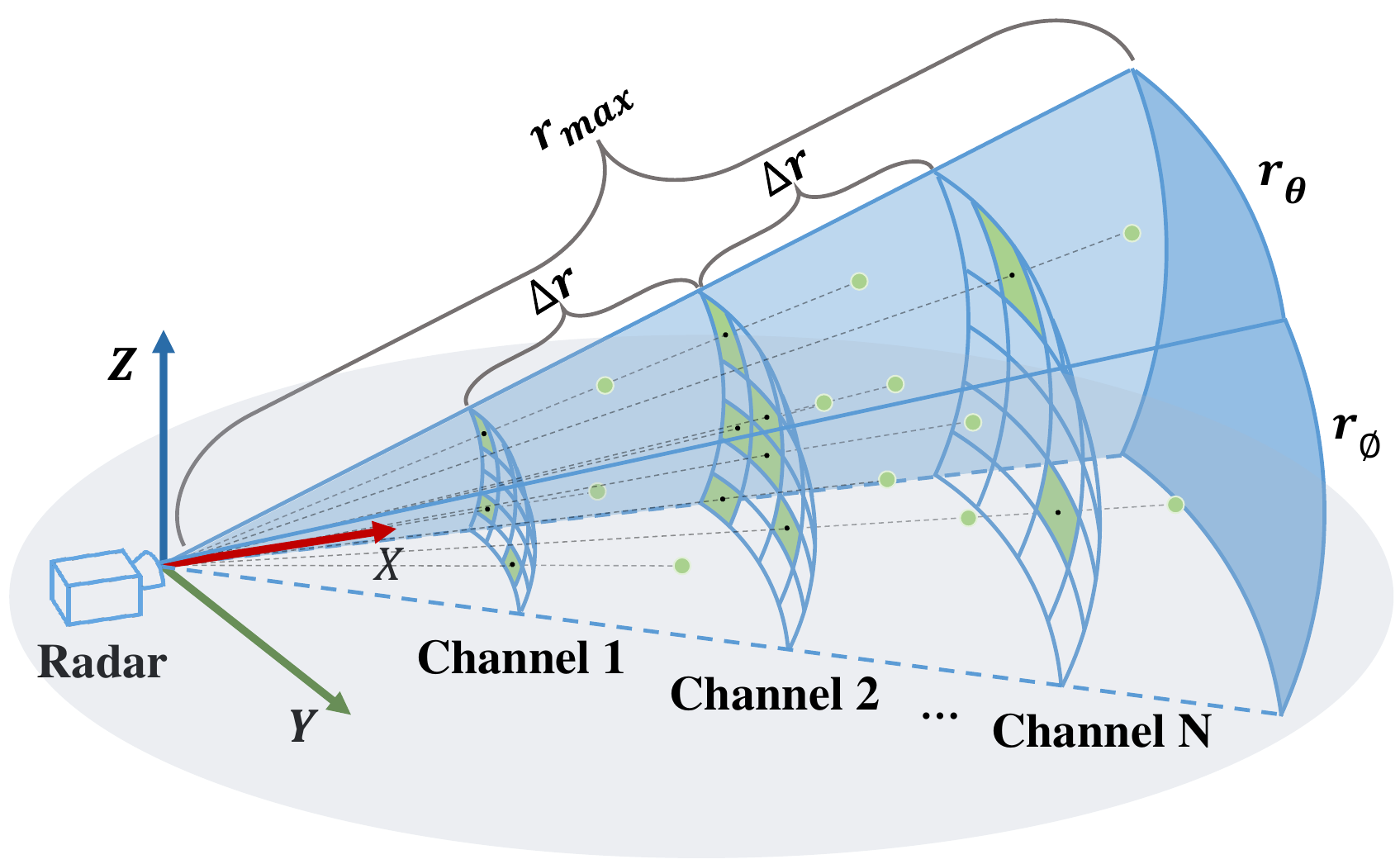}
        \caption{Illustration of multi-channel range image point cloud proxy representation for mmWave Radar.}
	\label{range_image_process}
\end{figure}

\section{Methods}
This section specifically details the proposed method. In recent years, image diffusion models, trained on massive image pairs, have significantly advanced the field of image super-resolution, inspiring us to leverage research from this domain to tackle point cloud super-resolution tasks. Consequently, we integrate range images with diffusion models. Specifically, the range image projected from the LiDAR point cloud is used to supervise the training of the diffusion model, after which high-quality range images are restored from noise, conditioned by range data from mmWave radar. Ultimately, the range image is back-projected to generate high-quality mmWave radar point clouds. The system architecture is illustrated in Fig. \ref{system}.

\subsection{Range Image Construction}
Certain LiDAR systems, such as Velodyne, generate raw data in a format that resembles range images. Each column represents the distances measured by laser range-finders at a specific moment, while each row corresponds to varying rotational angles of the sensors. This implies that range images can serve as a proxy for point clouds. A range image, in essence, is a 2D array where each pixel contains the spherical coordinates and range of a point mapped onto its field of view.
We transform each point $\Pi: \mathbb{R}^3 \rightarrow \mathbb{R}^2$ through the mapping $\mathbf{p_i}=(p_x, p_y, p_z)$ into spherical coordinates, and then into image coordinates, as follows:
\begin{equation}
    \label{range_image}
    \left( \begin{matrix}
     i_\theta\\
     i_\phi
    \end{matrix} \right) =
    \left( \begin{matrix}
     \lfloor \left( {\arctan 2(p_y, p_x)-\theta_{\text{min}}} \right)l_w {r_\theta}^{-1}\rfloor\\
     \lfloor \left( {\arccos (p_z, r)-\phi_{\text{min}}} \right)l_h {r_\phi}^{-1}\rfloor
    \end{matrix} \right),
\end{equation}
where $r=\|\mathbf{p_i}\|_2$ denotes the range of each point, $\theta_{\text {min }}$ and $\phi_{\text {min }}$ represent the minimum values of the azimuth and elevation angles, $r_\theta$ and $r_\phi$ are the ranges of the azimuth and elevation angles, $l_w$ and $l_h$ are the width and height of the range image, and $\lfloor \cdot \rfloor$ denotes the floor function.

\begin{figure*}[t]
    \begin{center}
        \includegraphics[width=\linewidth]{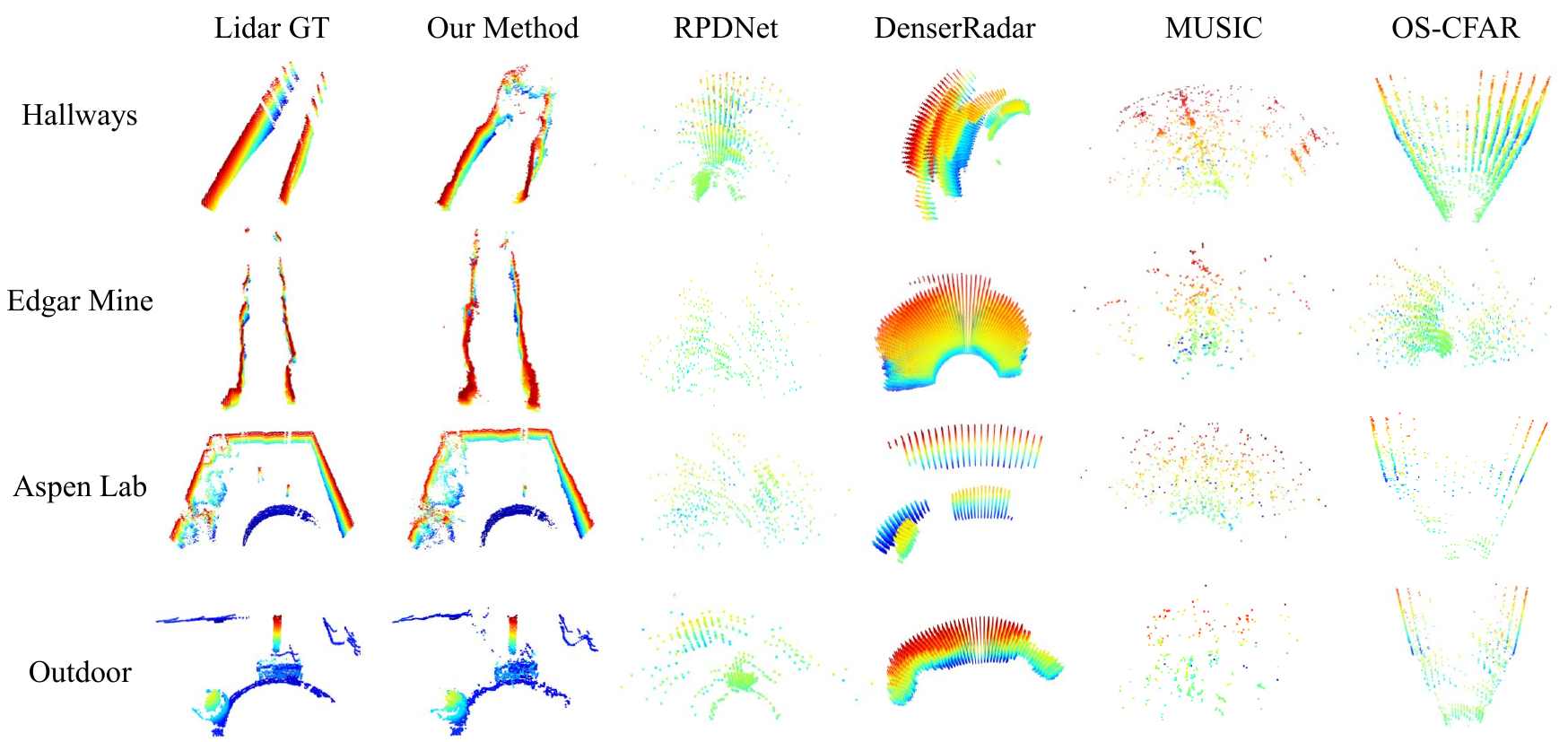}
    \end{center}
    \caption{Qualitative comparisons of single-frame 3D point clouds conducted on the ColoRadar dataset. The LiDAR ground truth and the 3D mmWave radar point clouds generated by each method in the Hallway, Edgar Mine, Aspen Lab, and Outdoor scenes are shown above. A multi-color rendering scheme is employed for visualizing the point clouds, with red hues representing higher elevations.}
    \label{experience}
\end{figure*}

\subsection{Data Representation}
% 多通道距离图像
Furthermore, given the inherent penetration ability of mmWave radar, directly converting the raw mmWave radar point cloud data into range images results in multiple points being projected into the conical region corresponding to each pixel. This phenomenon causes point cloud data at greater radial distances to become occluded, leading to data loss and increased sparsity of the mmWave radar data. To mitigate this, we perform slicing operations along the range dimension of the mmWave radar point cloud and individually map the radar data from each slice, thus generating multi-channel range images as the conditions of the diffusion model. The necessity of multiple channels is also validated in Section \ref{Range Image Channel Number}. The process of constructing multi-channel range images is illustrated in Fig. \ref{range_image_process}.

% 距离图像分辨率
Notably, the resolution of the range image is critically important. If the resolution is too low, significant amounts of critical information are lost during the point cloud mapping process. Conversely, excessively high resolution can result in an overly complex model architecture, placing undue computational strain and complicating subsequent training. For mmWave radar, which produces relatively unordered point cloud data, some information loss is inevitable during the mapping to range images. To strike a balance between model capacity and information retention, we carefully assess the amount of preserved information at various range image resolutions, ultimately selecting an optimal resolution. In contrast, the raw LiDAR point cloud, with its angular resolution and relatively organized spatial distribution, allows for the conversion to range images with minimal loss when a resolution approximating the LiDAR’s angular resolution is chosen.

\subsection{Diffusion-Based Range Image Prediction}
As elaborated in Section \ref{diffusion}, the principal objective of the diffusion model is to learn the reverse process of the forward diffusion mechanism, which incrementally introduces noise to data samples according to the predefined noise schedule described in Eq. \ref{diffusion2}, where $s(t)=1$ and $\sigma(t)=t$. Given a set of LiDAR range images $x_0$ and multi-channel radar range images $c$ that are aligned both spatially and temporally, Gaussian noise is introduced and propagated to $x_t$. We follow the parameterization approach for the reverse process proposed by Karras et al. \cite{karras2022elucidating} to model the original data sample $D_\theta(x_t, t, c)$ conditioned on $c$. During the inference stage, we perform deterministic sampling using the probabilistic flow (PF) ODE defined by Eq. \ref{diffusion1} to accelerate the inference process. The Heun method is employed for iterative solving. The PF ODE is defined as follows:

\begin{equation}
	\label{diffusion7}
	dx = -{\dot{\sigma}(t)}{\sigma(t)} \nabla_x \log p(x; \sigma(t))dt,
\end{equation} 
where the dot denotes a time derivative, $\nabla_x \log p(x; \sigma(t))$ represents the score function.

\subsubsection{Design of the Network Architecture}The resolutions of range images from LiDAR and mmWave radar differ due to distinct point cloud densities. If the range images of the mmWave radar are directly concatenated with the noise at the input layer, size alignment through methods such as interpolation becomes necessary, leading to the inevitable loss of information. Consequently, we opt to embed conditions during the downsampling stage, thereby ensuring the maximum retention of the original data. Furthermore, as the pitch range of LiDAR is narrower than its horizontal range, the height and width of its range images are unequal. Thus, we introduce a horizontal sampling module, which effectively extracts the lateral features of the target, adjusts the feature dimensions, and accelerates the feature extraction process simultaneously.
\begin{table*}[ht]
\renewcommand{\arraystretch}{1.4}
    \setlength{\abovecaptionskip}{4pt}
    \setlength{\belowcaptionskip}{4pt} 
    \caption{Quantitative results on the ColoRadar dataset. The \textbf{bold} denotes the best performance.}
    \label{Coloradar}
    \setlength{\belowcaptionskip}{4pt} 
    \centering
    \setlength{\tabcolsep}{1pt}
    \resizebox{\textwidth}{!}{
    \begin{tabular}{c|cccccccccccc}
        \Xhline{1.2pt}
        \multirow{2}{*}{Method} & \multicolumn{3}{c}{Hallways} & \multicolumn{3}{c}{Edgar Mine} & \multicolumn{3}{c}{Aspen Lab} & \multicolumn{3}{c}{Outdoor} \\
        \Xcline{2-13}{0.4pt}
        % \cmidrule(r){0-1}\cmidrule(lr){2-3}\cmidrule(lr){4-5}\cmidrule(lr){6-7}
         & CD(m) $\downarrow$ & MHD(m) $\downarrow$ & F-Score($\%$) $\uparrow$ & CD(m) $\downarrow$ & MHD(m) $\downarrow$ & F-Score($\%$) $\uparrow$ & CD(m) $\downarrow$ & MHD(m) $\downarrow$ & F-Score($\%$) $\uparrow$ & CD(m) $\downarrow$ & MHD(m) $\downarrow$ & F-Score($\%$) $\uparrow$ \\
        \Xhline{0.4pt}
        
        OS-CFAR \cite{rohling1983radar} & 5.741 & 0.902 & 20.8 & 8.327 & 0.963 & 20.8 & 4.79 & 1.064 & 25.6 & 9.20  & 2.459 & 18.9 \\
        MUSIC \cite{Schmidt1986Multiple} &  11.782 & 0.722 & 4.574 & 18.277 & 0.694 & 4.006 &  17.605 & 1.054 & 1.6 & 13.653 & 2.038 & 4.109 \\
        RPDNet \cite{cheng2022novel} &  2.329 & 0.545 & 32.8 & 3.044 & \textbf{0.464} & 33.7 & 2.927 & 0.858 & 30.7 & 8.183 & 2.203 & 24.3\\
        DenserRadar \cite{han2024denserradar} &  8.345 & 4.081 & 14.8 & 3.603 & 0.653 & 30.7 & 2.315 & 0.742 & 37.4 & 20.2 & 11.5 & 15.7\\
        
        \Xhline{0.4pt}

        Ours & \textbf{2.259} & \textbf{0.522} & \textbf{48.6} & \textbf{2.754} & 0.644 & \textbf{36.3} & \textbf{2.042} & \textbf{0.676} & \textbf{48.8} & \textbf{6.615} & \textbf{1.948} & \textbf{26}\\
 
        \Xhline{1.2pt}
    \end{tabular} 
    }
\end{table*}

\begin{table}[ht]
\renewcommand{\arraystretch}{1.4}
    \centering
    \setlength{\abovecaptionskip}{4pt}
    \setlength{\belowcaptionskip}{4pt} 
    \caption{Generalization testing of the Proposed Method.}

    \label{self-constructed}
    \setlength{\tabcolsep}{1pt}
    \resizebox{\linewidth}{!}{
    \begin{tabular}{c|cccccc}
        \Xhline{1.2pt}
        \multirow{2}{*}{Method} & \multicolumn{3}{c}{Corridor} & \multicolumn{3}{c}{Hall} \\
        \Xcline{2-7}{0.4pt}
        % \cmidrule(r){0-1}\cmidrule(lr){2-3}\cmidrule(lr){4-5}\cmidrule(lr){6-7}
         & CD(m) $\downarrow$ & MHD(m) $\downarrow$ & F-Score($\%$) $\uparrow$ & CD(m) $\downarrow$ & MHD(m) $\downarrow$ & F-Score($\%$) $\uparrow$ \\
        \Xhline{0.4pt}

        OS-CFAR \cite{rohling1983radar} & 29.712 & 3.08 & 2.1 & 17.294 & 3.713 & 2.5 \\
        Ours & \textbf{7.615} & \textbf{1.058} & \textbf{11.4} & \textbf{8.835} & \textbf{2.584} & \textbf{7.6} \\
        \Xhline{1.2pt}
    \end{tabular} }
\end{table}

\subsubsection{Training Objective}
We first optimize the following Mean Squared Error (MSE) loss to make $D_\theta(x_t, t, c)$ similar to $x_0$. The MSE loss is expressed as follows:

\begin{equation}
	\label{L_m}
	L_m=\left\|x_0-D_\theta\left(x_t, t, c\right)\right\|_2^2.
\end{equation}

However, since MSE loss is particularly sensitive to outliers, it may lead the model to overly focus on the few anomalous pixels within the image, thereby ignoring the broader enhancement in image quality. To mitigate this issue, we introduce the Learned Perceptual Image Patch Similarity (LPIPS) \cite{zhang2018unreasonable}, which leverages a pre-trained deep network to extract features from both $x_0$ and $D_\theta\left(x_t, t, c\right)$. and then calculates the distance between these features to evaluate the perceptual similarity between the images. The LPIPS loss is articulated as follows:

\begin{equation}
	\label{L_p}
	L_p=\left\|g_p\left(x_0\right)-g_p\left(D_\theta\left(x_t, t, c\right)\right)\right\|_2^2.
\end{equation}    

Furthermore, when transforming point clouds into range images, while the loss of geometric information remains negligible, it cannot be eliminated. To mitigate this geometric loss, drawing inspiration from the per-point coordinate supervision method in \cite{zhang2024towards}, we introduce per-pixel distance supervision. This involves calculating the distance loss for each pixel based on the predicted distance at each pixel location.

\begin{equation}
	\label{L_c}
	L_c=\left|x_0-D_\theta\left(x_t, t, c\right)\right|.
\end{equation} 

Finally, our loss function consists of three components.

\begin{equation}
	\label{L}
	L=\lambda_m L_m+\lambda_p L_p+\lambda_c L_c.
\end{equation}  
where $\lambda_m, \lambda_p, \lambda_c$ denote the weight coefficients of each loss function, used to balance multi-objective optimization during model training

\section{Experiments and Results}
In this section, we conduct benchmark comparisons, real-world experiments, and ablation experiments on both the public dataset (Coloradar) and the self-constructed dataset to validate the superior performance of our method.

\subsection{Benchmark Comparisons}
We performed benchmark comparisons with four baseline methods, including OS-CFAR \cite{rohling1983radar}, MUSIC \cite{Schmidt1986Multiple}, RPDNet \cite{cheng2022novel}, and DenserRadar \cite{han2024denserradar}. 

\subsubsection{Dataset Configuration}
The ColoRadar dataset provides LiDAR point cloud data and raw mmWave ADC data for seven different environments. Each environment contains multiple trajectories. We employ the data from the four scenes enumerated in Table \ref{Coloradar} for both training and testing. The first three trajectories of each scene constitute the training set, while the remaining trajectories are designated as the test set. To ensure a fair comparison, the dataset configurations of RPDNet \cite{cheng2022novel} and DenserRadar \cite{han2024denserradar} methods are aligned with those adopted in this study. 

\subsubsection{Dataset Preprocessing}
Due to the different operational ranges of mmWave radar and LiDAR, we preprocessed the raw sensor data. First, we used the extrinsic calibration matrix of the two sensor coordinate systems to achieve spatial alignment of the data. Subsequently, we obtained the shared field of view of the two sensors and removed data outside the range. In addition, due to the low reflectivity of surfaces such as floors and indoor ceilings, these objects are usually not perceived by millimeter wave radar. Therefore, we applied Patchwork++ \cite{lee2022patchwork++} to remove these point clouds from the LiDAR data, where the ceiling is removed by flipping the Z-axis and applying Patchwork++ again.

Additionally, considering that mmWave radar is not sensitive to low-reflectivity objects, we first employed DBSCAN to cluster the combined point cloud of mmWave radar and LiDAR. Then, we applied additional filtering to the LiDAR point cloud using labels from the mmWave radar to eliminate objects that are difficult for the mmWave radar to detect. This step ensures that the neural network does not learn features undetectable by mmWave radar, thereby preventing the neural network from being misled during training.

\begin{figure}[t]
	\centering
        \includegraphics[width=200pt]{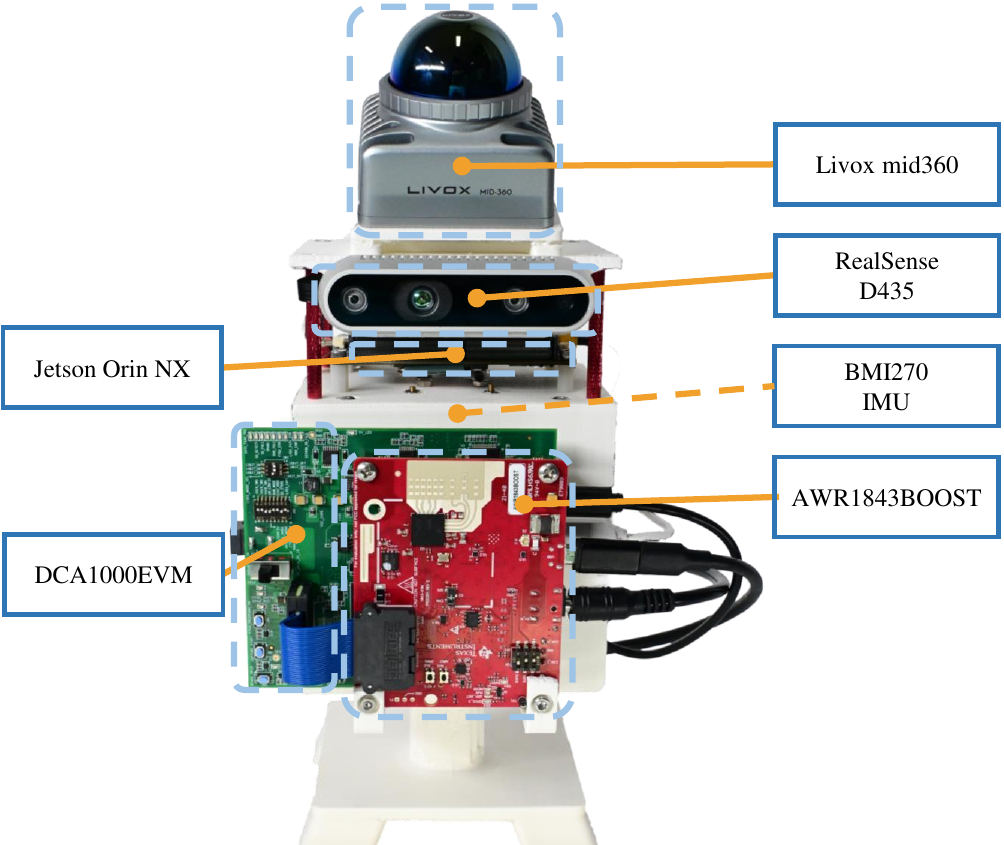}
        \caption{Our customized handheld data collection platform. The LiDAR is mounted at an inclined angle to align with the field of view of the mmWave radar.}
	\label{dataplatform}
\end{figure}

\subsubsection{Implementation}
% 叙述本文的CFAR的阈值设置
In the process of mmWave point cloud processing, we obtain the original mmWave radar point cloud data based on the CFAR. The performance of the CFAR detector is notably influenced by the threshold setting. A higher threshold enables superior denoising performance; however, it also leads to a loss of information. Under such conditions, it becomes nearly impossible to match the density of the mmWave radar point cloud to that of LiDAR point clouds. Therefore, to preserve as much raw mmWave radar information as possible, we configured the CFAR detector’s threshold to approach zero.

To balance model capacity and information, we set the range image resolution of LiDAR and radar to $128 \times 512$ and $64 \times 64$, respectively. We adopted the same diffusion noise and timestamp settings as the EDM proposed by Karras et al. \cite{karras2022elucidating}, and used the Heun deterministic sampler for sampling.

\subsubsection{Qualitative Comparison}
The mmWave radar point clouds generated by both the proposed and baseline methods, alongside the ground truth LiDAR point clouds, are presented in Fig. \ref{selfmade} and Fig. \ref{experience}. The results demonstrate that our method outperforms all baseline methods in terms of density across all scenes. Our approach produces high-fidelity scene details, reconstructs the curved geometry of objects, and maintains sharp corner features under complex environmental conditions. In contrast to the OS-CFAR \cite{rohling1983radar} and MUSIC \cite{Schmidt1986Multiple}, both RPDNet \cite{cheng2022novel} and DenserRadar \cite{han2024denserradar} employ deep networks as detectors and achieve significantly improved detection results. Nevertheless, the point clouds produced by these methods remain less dense and accurate than those generated by our proposed approach. Furthermore, DenserRadar \cite{han2024denserradar} operates directly on the voluminous raw radar data in the form of a 4-dimensional tensor (cube), which necessitates a model with substantially greater capacity.

\subsubsection{Quantitative Comparison}
\begin{figure}
    \centering
    \begin{subfigure}{0.48\linewidth}
        \centering
        \includegraphics[width=1\linewidth]{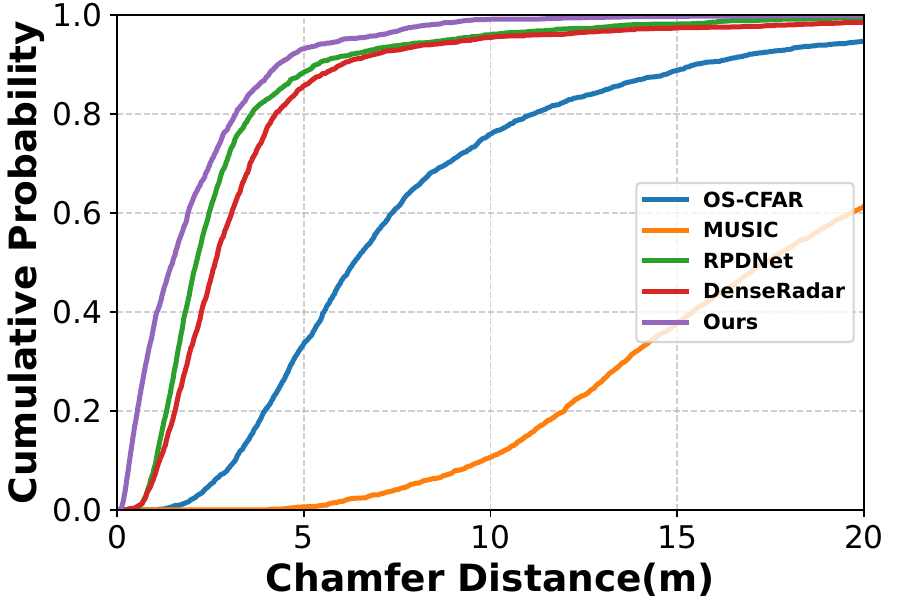}
        \caption{Hallways}
        \label{pic:hallways_cdf}
    \end{subfigure}
    \centering
    \begin{subfigure}{0.48\linewidth}
        \centering
        \includegraphics[width=1\linewidth]{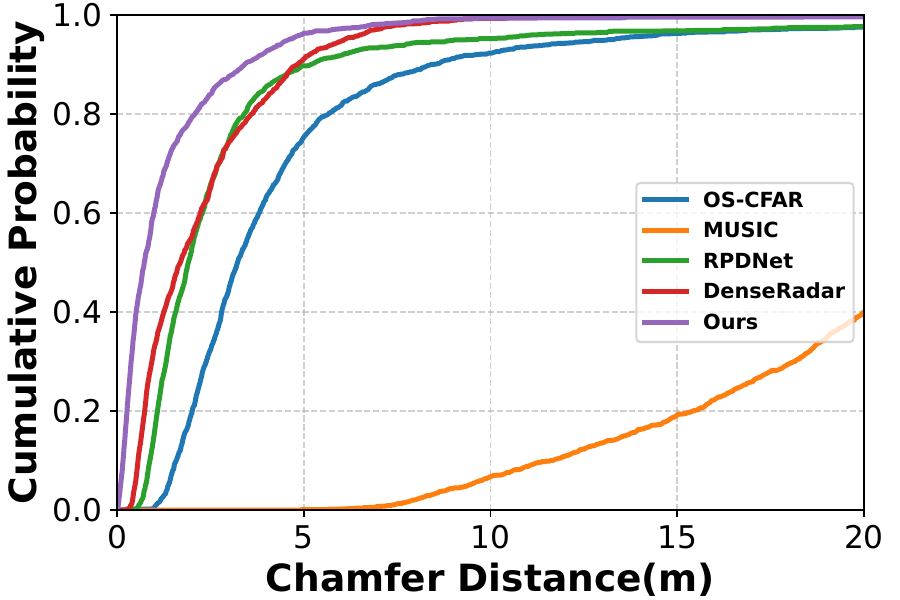}
        \caption{Edgar Mine}
        \label{pic:classroom_cdf}
    \end{subfigure}
    \centering
    \begin{subfigure}{0.48\linewidth}
        \centering
        \includegraphics[width=1\linewidth]{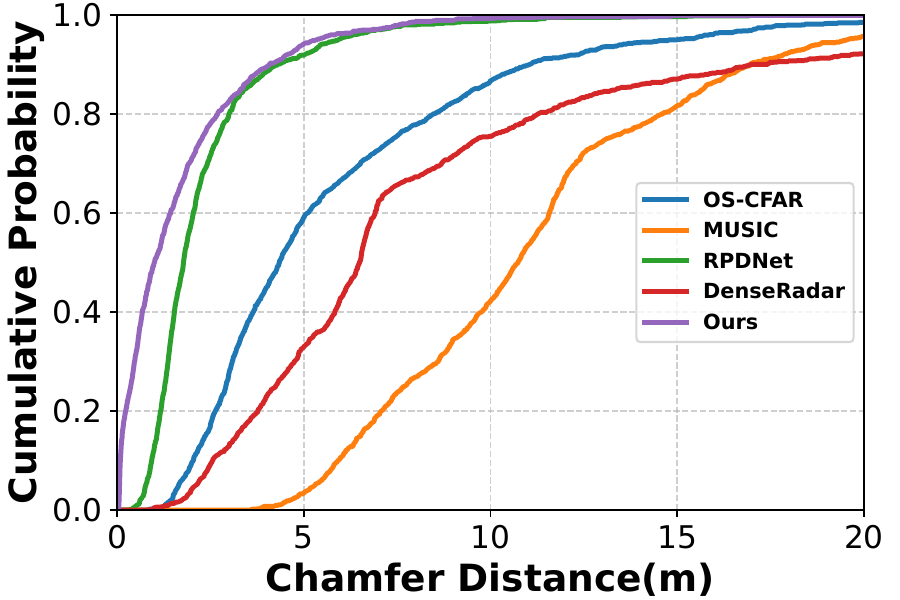}
        \caption{Aspen}
        \label{pic:aspen_cdf}
    \end{subfigure}
    \centering
    \begin{subfigure}{0.48\linewidth}
        \centering
        \includegraphics[width=1\linewidth]{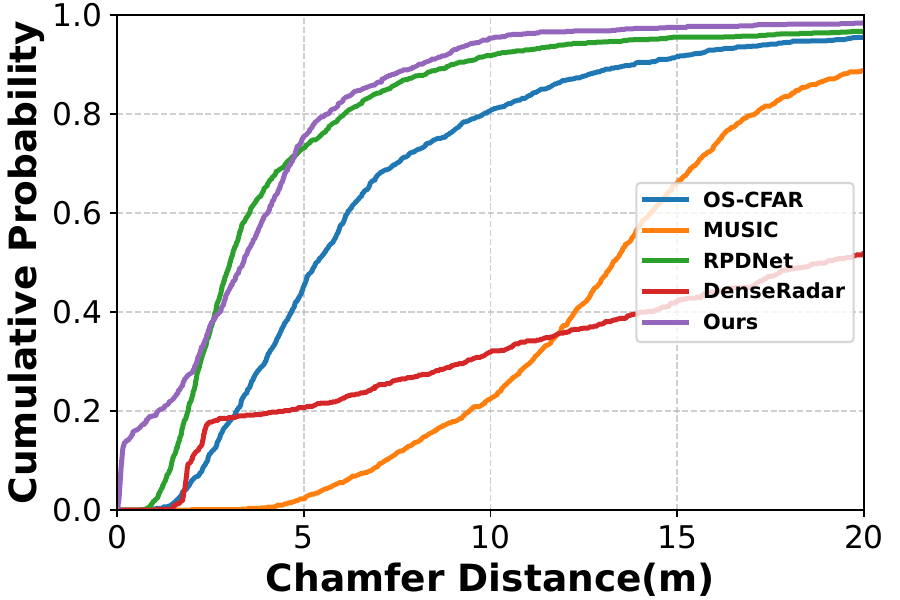}
        \caption{Outdoors}
        \label{pic:outdoors_cdf}
    \end{subfigure}
    \centering
    \caption{The CDF curves of our method and the baseline methods on the ColoRadar dataset.}
    \label{cdfs}
\end{figure}

We begin by presenting the evaluation metrics employed in our quantitative comparison: Chamfer Distance (CD), Modified Hausdorff Distance (MHD), and F-Score. CD quantifies the global similarity between two point clouds by summing the distances from each point to the nearest point in the other point cloud. MHD evaluates the disparity between two point clouds by calculating their forward and backward Hausdorff distances. F-score, defined as the harmonic mean of precision $P$ and recall $R$, is utilized to assess the quality of point cloud reconstruction. The results of the quantitative analysis are provided in Tab.\ref{Coloradar}, while Fig. \ref{cdfs} illustrates the cumulative distribution function (CDF) curves of the Chamfer Distance.

The results demonstrate that the proposed method significantly outperforms the baseline approaches across all scenes in the ColoRadar datasets. In the Edgar Mine scene of ColoRadar datasets, the RPDNet \cite{cheng2022novel} methodis only slightly better than our approach with respect to the MHD metric. This is primarily because the Edgar Mine scene is dominated by walls with simple semantics and geometry, which limits the advantage of our method that leverages rich semantic and geometric information.

\begin{table*}[ht]
\renewcommand{\arraystretch}{1.4}
    \setlength{\abovecaptionskip}{4pt}
    \setlength{\belowcaptionskip}{4pt} 
    \caption{Ablation Experiment of the Data Representation Format.}
    \label{ablation1}
    \setlength{\belowcaptionskip}{4pt} 
    \centering
    \setlength{\tabcolsep}{1pt}
    \resizebox{\textwidth}{!}{
    \begin{tabular}{c|cccccccccccc}
        \Xhline{1.2pt}
        \multirow{2}{*}{Method} & \multicolumn{3}{c}{Hallways} & \multicolumn{3}{c}{Edgar Mine} & \multicolumn{3}{c}{Aspen Lab} & \multicolumn{3}{c}{Outdoor} \\
        \Xcline{2-13}{0.4pt}
        % \cmidrule(r){0-1}\cmidrule(lr){2-3}\cmidrule(lr){4-5}\cmidrule(lr){6-7}
         & CD(m) $\downarrow$ & MHD(m) $\downarrow$ & F-Score($\%$) $\uparrow$ & CD(m) $\downarrow$ & MHD(m) $\downarrow$ & F-Score($\%$) $\uparrow$ & CD(m) $\downarrow$ & MHD(m) $\downarrow$ & F-Score($\%$) $\uparrow$ & CD(m) $\downarrow$ & MHD(m) $\downarrow$ & F-Score($\%$) $\uparrow$ \\
        \Xhline{0.4pt}
        
        Ours-BEV  & 2.452 & 0.548 & \textbf{51.808} & 5.127 & 1.154 & 30.374 & 2.462 & 0.723 & 47.234 & 20.155 & 8.358  & 18.113 \\
        Ours-RI & \textbf{2.259} & \textbf{0.522} & 48.6 & \textbf{2.754} & \textbf{0.644} & \textbf{36.3} & \textbf{2.042} & \textbf{0.676} & \textbf{48.8} & \textbf{6.615} & \textbf{1.948} & \textbf{26.0}\\
        
        \Xhline{1.2pt}
    \end{tabular} 
    }
{\scriptsize%
\textbf{Note:} Ours-RI denotes our method using Range Image as the data representation format, while Ours-BEV represents our approach with Bird's Eye View representation.
}
\end{table*}

\begin{figure}
    \centering
    \begin{subfigure}{0.48\linewidth}
        \centering
        \includegraphics[width=1\linewidth]{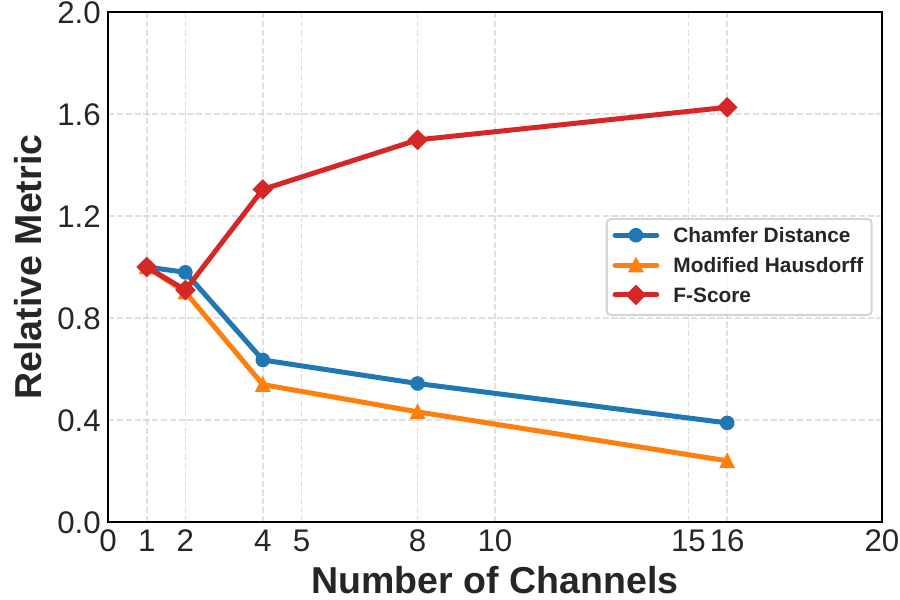}
        \caption{Hallways}
        \label{pic:hallways}
    \end{subfigure}
    \centering
    \begin{subfigure}{0.48\linewidth}
        \centering
        \includegraphics[width=1\linewidth]{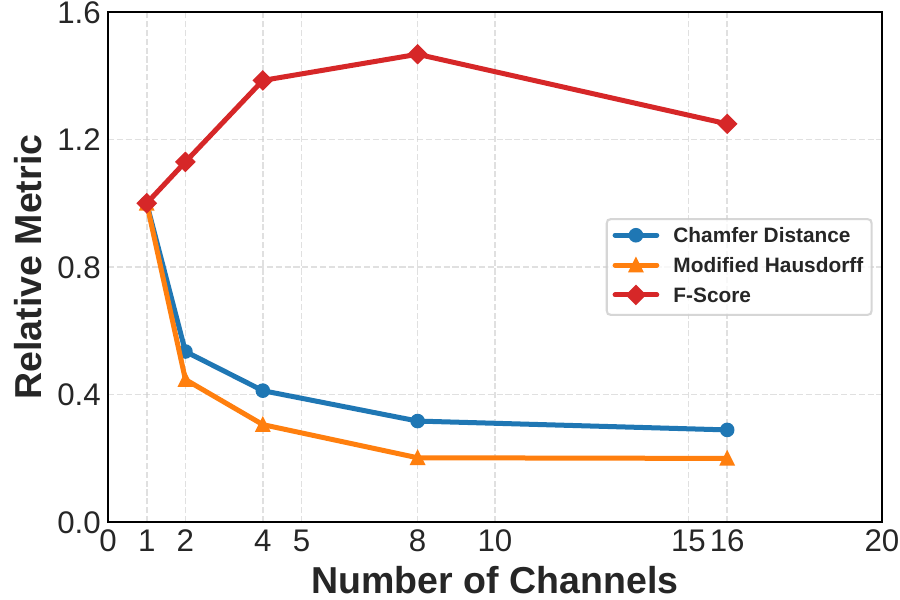}
        \caption{Edgar Mine}
        \label{pic:edgar mine}
    \end{subfigure}
    \centering
    \begin{subfigure}{0.48\linewidth}
        \centering
        \includegraphics[width=1\linewidth]{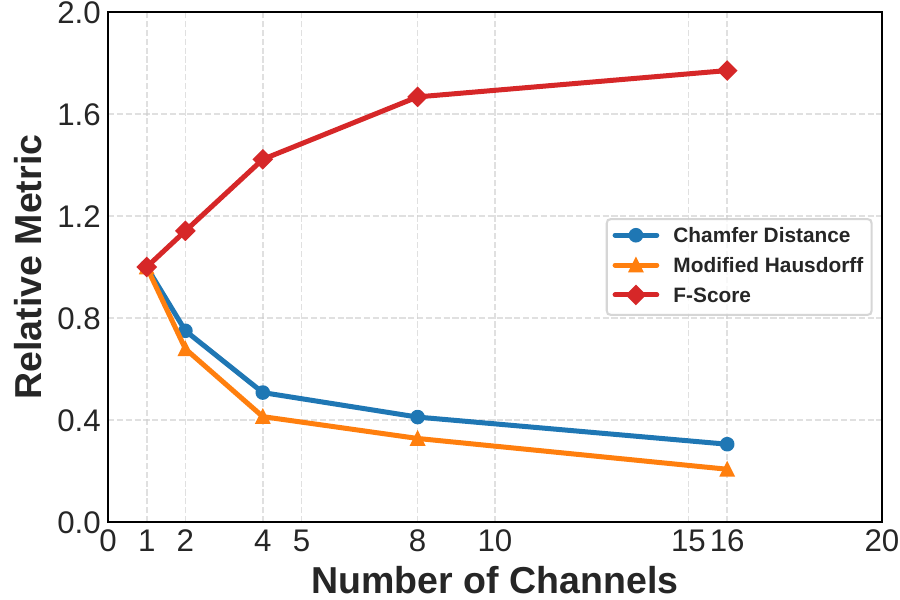}
        \caption{Aspen}
        \label{pic:aspen}
    \end{subfigure}
    \centering
    \begin{subfigure}{0.48\linewidth}
        \centering
        \includegraphics[width=1\linewidth]{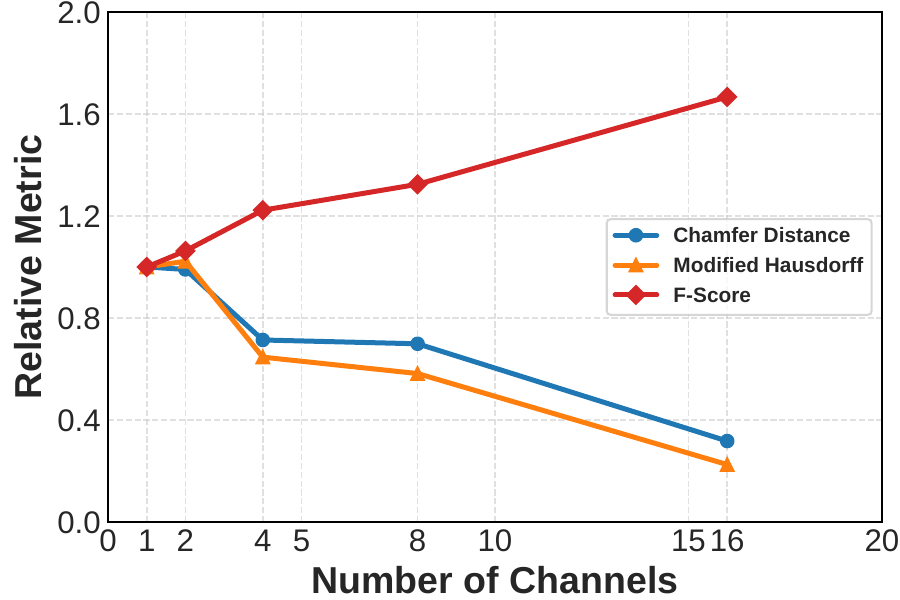}
        \caption{Outdoors}
        \label{pic:outdoors}
    \end{subfigure}
    \centering
    \caption{Ablation study on the number of channels in mmWave radar range images. The y-axis denotes the relative proportion of each metric with respect to the case when the number of channels is one.}
    \label{ablation2}
    % \vspace{-0.7cm}
\end{figure}

\subsection{Ablation Experiments}
Beyond benchmark comparisons, we conducted ablation experiments to substantiate the necessity of the proposed method.

\subsubsection{Data Representation Format}
In this experiment, the data representation in our method is switched from range images to BEV, a format widely used in prior studies \cite{zhang2024towards, Luan2024Diffusion, wu2024diffradar}, while all other components remain unchanged. Table \ref{ablation1} reports the quantitative results: representing the data as range images yields substantially higher point‑cloud quality than BEV. This suggests that range images, which align with human perceptual priors, more effectively exploit the capabilities of the diffusion model, thereby underscoring the efficacy of the proposed approach.

\subsubsection{Range Image Channel Number}
\label{Range Image Channel Number}
Furthermore, we performed an ablation study on the range image channels. Only the number of range image channels was altered in this experiment, and all other settings remained identical to the benchmark comparisons. The results are presented in Fig. \ref{ablation2}. It is immediately noticeable that all multi-channel range images substantially surpassed their single-channel counterparts in performance. This phenomenon stems from the intrinsic penetrative capability of mmWave radar, which produces numerous data points along identical trajectories. Employing solely single-channel range imagery culminates in multiple points being superimposed upon identical pixels, thereby precipitating substantial information loss. The experiment substantiates the imperative for employing multi-channel range images, with 16-channel configurations yielding better performance.

\subsection{Real-world Experiments}
We further assess the generalization capability of the proposed method by employing our self-constructed datasets.

\subsubsection{Data Collection Platform}
We customized a handheld data collection platform, as shown in Fig. \ref{dataplatform}. It consists of an NVIDIA Jetson Orin NX for recording sensor data, a Livox Mid-360 LiDAR, and a BMI270 IMU to estimate real-time platform poses with the assistance of Fast-LIO \cite{Xu2021FASTLIO}. Additionally, it is equipped with a TI-AWR1843BOOST and TI-DCA1000EVM for collecting mmWave radar raw data, and an Intel RealSense D435 to capture image data for visualization.Using the handheld data collection platform, we respectively collected four trajectories in the hall and corridor, totaling 6,636 frames.

\subsubsection{Generalization Capability Tests}
We deployed the model trained on the ColoRadar dataset directly on our self-collected dataset, without any further fine-tuning, and benchmarked it against CFAR-based \cite{rohling1983radar} baselines. This test has achieved robust generalization across heterogeneous scenes, LiDAR sensors, and radar parameterizations. Quantitative results are reported in Table \ref{self-constructed}, with metrics computed using LiDAR point clouds as ground truth. Even under unseen scenarios and sensor configurations, the proposed method substantially outperforms conventional CFAR-based approaches, compellingly attesting to its efficacy and generalization capacity.

\section{Conclusion}
This paper proposes a novel approach for generating high-quality mmWave radar point clouds. By fully leveraging the powerful generation capability of the diffusion model, the proposed method can effectively process the sparse and noisy original mmWave radar point cloud of a single frame and obtain a point cloud similar to LiDAR. Our approach employs range images aligned with human perception as proxy representations for point clouds and combines them with image diffusion models for the first time to solve this task. Owing to the projection that aligns with human perception, the range image representation closely resembles natural images, thereby facilitating the transfer of knowledge from the pre-trained image diffusion model and substantially enhancing overall performance. The proposed method is validated on public and self-constructed datasets, significantly outperforming baseline methods in terms of both point cloud quality and density.

In the future, we intend to explore the interdependencies between adjacent frames of mmWave radar data to further enhance the interframe continuity of mmWave radar point clouds. This is of great significance for applying mmWave radar to advanced tasks such as autonomous robot navigation.

\bibliographystyle{ieeetr}
\balance
\bibliography{paper}

\begin{thebibliography}{10}

\bibitem{lyu2024robust}
Y.~Lyu, L.~Hua, J.~Wu, X.~Liang, and C.~Zhao, ``Robust radar inertial odometry in dynamic 3d environments,'' {\em Drones}, vol.~8, no.~5, p.~197, 2024.

\bibitem{zhang20234dradarslam}
J.~Zhang, H.~Zhuge, Z.~Wu, G.~Peng, M.~Wen, Y.~Liu, and D.~Wang, ``4dradarslam: A 4d imaging radar slam system for large-scale environments based on pose graph optimization,'' in {\em 2023 IEEE International Conference on Robotics and Automation (ICRA)}, pp.~8333--8340, IEEE, 2023.

\bibitem{guan2023achelous}
R.~Guan, S.~Yao, X.~Zhu, K.~L. Man, E.~G. Lim, J.~Smith, Y.~Yue, and Y.~Yue, ``Achelous: A fast unified water-surface panoptic perception framework based on fusion of monocular camera and 4d mmwave radar,'' in {\em 2023 IEEE 26th International Conference on Intelligent Transportation Systems (ITSC)}, pp.~182--188, IEEE, 2023.

\bibitem{cheng2021robust}
Y.~Cheng, H.~Xu, and Y.~Liu, ``Robust small object detection on the water surface through fusion of camera and millimeter wave radar,'' in {\em Proceedings of the IEEE/CVF international conference on computer vision}, pp.~15263--15272, 2021.

\bibitem{cheng2022novel}
Y.~Cheng, J.~Su, M.~Jiang, and Y.~Liu, ``A novel radar point cloud generation method for robot environment perception,'' {\em IEEE Transactions on Robotics}, vol.~38, no.~6, pp.~3754--3773, 2022.

\bibitem{prabhakara2023high}
A.~Prabhakara, T.~Jin, A.~Das, G.~Bhatt, L.~Kumari, E.~Soltanaghai, J.~Bilmes, S.~Kumar, and A.~Rowe, ``High resolution point clouds from mmwave radar,'' in {\em 2023 IEEE International Conference on Robotics and Automation (ICRA)}, pp.~4135--4142, IEEE, 2023.

\bibitem{zhang2024towards}
R.~Zhang, D.~Xue, Y.~Wang, R.~Geng, and F.~Gao, ``Towards dense and accurate radar perception via efficient cross-modal diffusion model,'' {\em arXiv preprint arXiv:2403.08460}, 2024.

\bibitem{Luan2024Diffusion}
K.~Luan, C.~Shi, N.~Wang, Y.~Cheng, H.~Lu, and X.~Chen, ``Diffusion-based point cloud super-resolution for mmwave radar data,'' in {\em 2024 IEEE International Conference on Robotics and Automation (ICRA)}, pp.~11171--11177, 2024.

\bibitem{wu2024diffradar}
J.~Wu, R.~Geng, Y.~Li, D.~Zhang, Z.~Lu, Y.~Hu, and Y.~Chen, ``Diffradar: high-quality mmwave radar perception with diffusion probabilistic model,'' in {\em ICASSP 2024-2024 IEEE International Conference on Acoustics, Speech and Signal Processing (ICASSP)}, pp.~8291--8295, IEEE, 2024.

\bibitem{barkat1989cfar}
M.~Barkat, S.~Himonas, and P.~Varshney, ``Cfar detection for multiple target situations,'' in {\em IEE Proceedings F (Radar and Signal Processing)}, vol.~136, pp.~193--209, IET, 1989.

\bibitem{minkler1990cfar}
G.~Minkler and J.~Minkler, ``Cfar: the principles of automatic radar detection in clutter,'' {\em Nasa sti/recon technical report a}, vol.~90, p.~23371, 1990.

\bibitem{gandhi1988analysis}
P.~P. Gandhi and S.~A. Kassam, ``Analysis of cfar processors in nonhomogeneous background,'' {\em IEEE Transactions on Aerospace and Electronic systems}, vol.~24, no.~4, pp.~427--445, 1988.

\bibitem{rohling1983radar}
H.~Rohling, ``Radar cfar thresholding in clutter and multiple target situations,'' {\em IEEE transactions on aerospace and electronic systems}, no.~4, pp.~608--621, 1983.

\bibitem{Schmidt1986Multiple}
R.~Schmidt, ``Multiple emitter location and signal parameter estimation,'' {\em IEEE Transactions on Antennas and Propagation}, vol.~34, no.~3, pp.~276--280, 1986.

\bibitem{han2024denserradar}
Z.~Han, J.~Jiang, X.~Ding, J.~Wang, Q.~Meng, S.~Xu, L.~He, and J.~Wang, ``Denserradar: A 4d millimeter-wave radar point cloud detector based on dense lidar point clouds,'' in {\em 2024 IEEE 27th International Conference on Intelligent Transportation Systems (ITSC)}, pp.~930--936, IEEE, 2024.

\bibitem{fan2024enhancing}
C.~Fan, S.~Zhang, K.~Liu, S.~Wang, Z.~Yang, and W.~Wang, ``Enhancing mmwave radar point cloud via visual-inertial supervision,'' in {\em 2024 IEEE International Conference on Robotics and Automation (ICRA)}, pp.~9010--9017, IEEE, 2024.

\bibitem{lu2020see}
C.~X. Lu, S.~Rosa, P.~Zhao, B.~Wang, C.~Chen, J.~A. Stankovic, N.~Trigoni, and A.~Markham, ``See through smoke: robust indoor mapping with low-cost mmwave radar,'' in {\em Proceedings of the 18th International Conference on Mobile Systems, Applications, and Services}, pp.~14--27, 2020.

\bibitem{geng2023dream}
R.~Geng, Y.~Li, D.~Zhang, J.~Wu, Y.~Gao, Y.~Hu, and Y.~Chen, ``Dream-pcd: Deep reconstruction and enhancement of mmwave radar pointcloud,'' {\em arXiv preprint arXiv:2309.15374}, 2023.

\bibitem{cai2023millipcd}
P.~Cai and S.~Sur, ``Millipcd: Beyond traditional vision indoor point cloud generation via handheld millimeter-wave devices,'' {\em Proceedings of the ACM on Interactive, Mobile, Wearable and Ubiquitous Technologies}, vol.~6, no.~4, pp.~1--24, 2023.

\bibitem{karras2022elucidating}
T.~Karras, M.~Aittala, T.~Aila, and S.~Laine, ``Elucidating the design space of diffusion-based generative models,'' {\em Advances in neural information processing systems}, vol.~35, pp.~26565--26577, 2022.

\bibitem{zhang2018unreasonable}
R.~Zhang, P.~Isola, A.~A. Efros, E.~Shechtman, and O.~Wang, ``The unreasonable effectiveness of deep features as a perceptual metric,'' in {\em Proceedings of the IEEE conference on computer vision and pattern recognition}, pp.~586--595, 2018.

\bibitem{lee2022patchwork++}
S.~Lee, H.~Lim, and H.~Myung, ``Patchwork++: Fast and robust ground segmentation solving partial under-segmentation using 3d point cloud,'' in {\em 2022 IEEE/RSJ International Conference on Intelligent Robots and Systems (IROS)}, pp.~13276--13283, IEEE, 2022.

\bibitem{Xu2021FASTLIO}
W.~Xu and F.~Zhang, ``Fast-lio: A fast, robust lidar-inertial odometry package by tightly-coupled iterated kalman filter,'' {\em IEEE Robotics and Automation Letters}, vol.~6, no.~2, pp.~3317--3324, 2021.

\end{thebibliography}

\end{document}